\documentclass{article}

\usepackage{microtype}
\usepackage{graphicx}
\usepackage{subcaption}
\usepackage{booktabs} 

\usepackage{hyperref}
\usepackage{xurl}

\usepackage[accepted]{icml2026}

\usepackage{amsmath}
\usepackage{amssymb}
\usepackage{mathtools}
\usepackage{amsthm}

\usepackage[capitalize,noabbrev]{cleveref}

\usepackage{tikz}
\usetikzlibrary{calc}
\usepackage{pgffor}
\usepackage{listings}
\usepackage[most]{tcolorbox}
\usepackage[export]{adjustbox}
\usepackage{placeins}
\usepackage[abs]{overpic}
\usepackage{xcolor}
\usepackage{enumitem}
\usepackage{longtable}
\usepackage{float}

\newcommand{\PatientRecord}[1]{%
    \begin{tikzpicture}[baseline=(txt.base), scale=0.64, transform shape, every node/.append style = {transform shape}]
        \def\foldsize{3.5pt}
        \def\padding{3pt}
        \def\iconwidth{1.2em}
        
        \node[font=\bfseries\sffamily\small, inner sep=0pt, anchor=west, opacity=0] (txt) at (\iconwidth, 0) {#1};
        
        \coordinate (nw) at ($(txt.north west) + (-\iconwidth - \padding, \padding)$);
        \coordinate (sw) at ($(txt.south west) + (-\iconwidth - \padding, -\padding)$);
        \coordinate (se) at ($(txt.south east) + (\padding, -\padding)$);
        \coordinate (ne) at ($(txt.north east) + (\padding, \padding)$);
        
        \draw[thick, draw=black!80, fill=white] 
            (sw) -- (nw) -- 
            ($(ne) - (\foldsize, 0)$) -- 
            ($(ne) - (0, \foldsize)$) -- 
            (se) -- cycle;
            
        \draw[thick, draw=black!80, line join=bevel] 
            ($(ne) - (\foldsize, 0)$) |- ($(ne) - (0, \foldsize)$) -- cycle;
            
        \coordinate (iconcenter) at ($(sw)!0.5!(nw) + (0.5*\iconwidth + 0.5*\padding, 0)$);
        
        \draw[red, line width=1.5pt, line cap=rect] 
            ($(iconcenter) - (2.5pt, 0)$) -- ($(iconcenter) + (2.5pt, 0)$); 
        \draw[red, line width=1.5pt, line cap=rect] 
            ($(iconcenter) - (0, 2.5pt)$) -- ($(iconcenter) + (0, 2.5pt)$);
            
        \node[font=\bfseries\sffamily\small, inner sep=0pt, anchor=center] at (txt.center) {#1};
            
    \end{tikzpicture}%
}

\newtcolorbox{questioncard}[1][]{
    enhanced,
    colback=white,
    colframe=gray,
    coltitle=black,
    fonttitle=\footnotesize\itshape,
    arc=3pt,
    boxsep=3pt,
    boxrule=0.5pt,
    left=2pt, right=2pt, bottom=2pt,
    top=1pt, 
    title={#1},
    attach boxed title to top center={yshift*=-3mm},
    boxed title style={colback=white, colframe=white, arc=3pt, boxsep=2pt}
}

\newcommand{\PopMedSeparator}{%
    \\
    \noindent{\color{gray!50}\hrulefill}
    \par%
}

\newcommand{\PopMedAnswerStart}{
    \noindent\textit{\textrm{Answer:}}
}

\newcommand{\PopMedBool}[1]{%
    \texttt{#1}%
}

\newcommand{\PopMedListWrapper}[1]{%
    \texttt{[#1]}%
}

\newcommand{\PopMedListJoin}{%
    \texttt{,}\allowbreak%
}

\newcommand{\PopMedDictWrapper}[1]{%
    \texttt{\{#1\}}%
}

\newcommand{\PopMedDictPair}[2]{%
    #1\texttt{:}#2%
}

\newcommand{\popmedquestion}[1]{%
\setlength{\parskip}{0.5\baselineskip}
\fontsize{7pt}{8.5pt}\selectfont
    \IfFileExists{questions/#1.tex}{\input{questions/#1.tex}}{}%
}

\definecolor{commentred}{RGB}{226,67,83}
\newcommand{\promptexample}[2]{%
  \lstinputlisting[linewidth=#2\textwidth]{prompt-examples/#1.txt}%
}

\theoremstyle{plain}
\newtheorem{theorem}{Theorem}[section]

\theoremstyle{definition}
\newtheorem{definition}[theorem]{Definition}

\theoremstyle{remark}

\icmltitlerunning{The Verbose Context Problem in Medical Records}

\begin{document}

\twocolumn[
\icmltitle{The Verbose Context Problem in Medical Records}

  \begin{icmlauthorlist}
    \icmlauthor{Shiva Kaul}{nothphci}
    \icmlauthor{Min-Gyu Kim}{ajou}
    \icmlauthor{Anjum Khurshid}{hphci}
    \icmlauthor{Sriram Vishwanath}{gatech}
  \end{icmlauthorlist}

  \icmlaffiliation{nothphci}{Work completed at Department of Population Medicine prior to current affiliation.}
  \icmlaffiliation{ajou}{Department of Biomedical Informatics, Ajou University School of Medicine}
  \icmlaffiliation{hphci}{Department of Population Medicine, Harvard Pilgrim Health Care Institute and Harvard Medical School}
  \icmlaffiliation{gatech}{School of Electrical and Computer Engineering, Georgia Institute of Technology}

  \icmlcorrespondingauthor{Shiva Kaul}{me@shivakaul.com}

\icmlkeywords{Machine Learning, ICML, Health, Long-Context}

\vskip 0.3in
]

\printAffiliationsAndNotice{}

\begin{abstract}
The verbose context problem occurs when structured concepts have token-inefficient textual representations. This bottleneck is acute in population health: cohort-level analysis of longitudinal patient records requires reasoning over thousands of medically-coded events, often exceeding 400K tokens. We present \href{https://shivakaul.com/x/popmedqa}{PopMedQA}, a benchmark isolating this problem through computational tasks on groups of longitudinal patient records. We construct the benchmark using \href{https://shivakaul.com/x/neopatient}{neopatient}, a new library for language-controlled generation of artificial patient records. Through extensive ablations—including prompting strategies, prompt compression, and agentic decomposition—we find that domain-independent methods fail to alleviate the verbose context problem. There remains significant opportunity to exploit domain-specific structure in language model inputs for population-scale reasoning.
\end{abstract}

\section{Introduction}
\label{sec:introduction}

\begin{figure*}[t!]
\noindent
\begin{tcbraster}[raster columns=3, raster width=\linewidth, raster equal height, raster column skip=1mm, raster force size]
\begin{questioncard}[Clustering]
\input{questions/big3-clustering.tex}
\end{questioncard} \begin{questioncard}[Top-$k$]
\input{questions/big3-topk.tex}
\end{questioncard} \begin{questioncard}[Two-Sample Test]
\input{questions/big3-twosample.tex}
\end{questioncard}
\end{tcbraster}

\includegraphics[width=\linewidth]{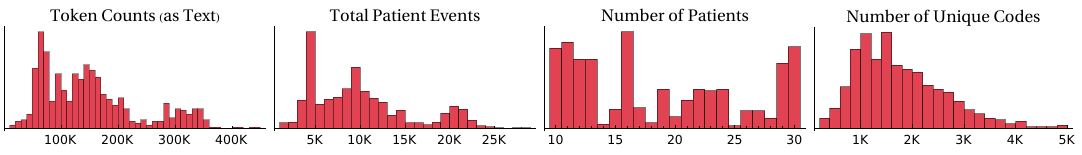}
\caption{Example questions from PopMedQA (above), as well as statistics of all its questions (below). Within the questions, each patient record is visually depicted as a boxed patient ID number. Since these records may contains thousands of coded events, and codes have verbose textual representations, PopMedQA is a long-context benchmark.}
\label{fig:popmed_combined}
\end{figure*}

Population health analytics focuses on identifying patterns, detecting anomalies, and quantifying disease burden across large groups of individuals. While large language models (LLMs) offer a flexible alternative to traditional rule-based risk adjustment systems, their application is hindered by the \emph{verbose context problem}. This problem arises when structured concepts—such as medical codes in electronic health records (EHRs)—have token-inefficient textual representations that inflate context lengths. For example, ``ICD-10 I21: Acute myocardial infarction'' is a textual representation $\textrm{str}(c)$ of the underlying concept $c$ of a heart attack. The usual process is to tokenize the string, lookup the token embeddings, and process the resulting sequence of vectors by the model $f$. In this notation, the language model's output is: $f \circ \mathrm{emb} \circ \mathrm{tok} \circ \mathrm{str} (c) \coloneqq y$. When such concepts pervade the context, as they do in longitudinal patient records, the total context length becomes too long for effective, efficient reasoning.

We capture the verbose context problem in a new benchmark called PopMedQA. Each of its questions involves reasoning over groups of 10-50 synthetic longitudinal patient records, which typically amount to 64K-256K tokens in a textual representation. As described in \Cref{sec:popmedqa_main} and \Cref{sec:verbosecontext}, it differs from existing long-context benchmarks in two ways. First, it specifically isolates how verbosity, rather than the presence of irrelevant ``hay'' context, affects performance. Second, it involves multi-hop reasoning over population-scale cohorts that cannot be decomposed into questions about individuals. Constructed with clinician and expert review, PopMedQA reflects the real-world priorities of population health, such as identifying latent clinical clusters or detecting sparse anomalies across disparate patient trajectories.

To construct the large volume of synthetic data required for PopMedQA, we introduce neopatient, a new software library for language-controlled generation of artificial patient records. Unlike rule-based generators (like Synthea~\cite{walonoski2018synthea}), neopatient trajectories are controlled through natural language descriptions, allowing for the creation of complex clinical cohorts without custom simulation code. Details on neopatient are provided in \Cref{sec:neopatient}.

In \Cref{sec:experiments-popmedqa}, we conduct a thorough evaluation of a range of language models on PopMedQA. These confirm the claimed design characteristics of PopMedQA, such as decomposition resistance. For ablations, we conduct a meta-analysis of multiple families of techniques for improving long-context performance, including prompting strategies, prompt compression, and agentic decomposition. 

Our analysis reveals several generalizable insights on the nature of population-scale EHR reasoning: 
(1) both clinical competence and long-context capability are required; 
(2) generic prompt compression is fragile;
(3) medical pretraining does not substantially improve performance; and
(4) agentic decomposition is ineffective and/or cost-prohibitive.  
Overall, we find that domain-independent methods fail to alleviate the verbose context problem, exposing a significant unrealized opportunity to exploit domain structure for population-scale reasoning.

\section{PopMedQA: Verbosity in Medical Records}
\label{sec:popmedqa_main}

To embody the verbose context problem, we present a new benchmark. PopMedQA consists of questions about groups (of size between 10 and 50) of longitudinal patient records. These records are drawn from over 25 thousand synthetic patients generated specifically for the benchmark. These records comprise over 14M medically-coded events.  When a typical code is conveyed as text, it uses between 8 and 20 tokens. A code accompanies each timestamped event. There are typically thousands of such events in each record. This makes most of PopMedQA's questions approximately 64K, 128K, or 256K tokens in length, when represented as text.

An important design feature of PopMedQA is \emph{decomposition resistance}. Long context can often be circumvented by partitioning it into separate chunks, process the chunks separately, and aggregating an answer from multiple rounds of processing. The purpose of PopMedQA is to more specifically reward verbose-context capability rather than this more routine context engineering. Accordingly, most of PopMedQA's questions are designed so that they cannot be readily solved by asking a series of individual patient-level questions. Each question poses one of nine computational tasks (see \Cref{fig:popmed_all_tasks,appx:task_scoring}). Some of these, such as planted clique and clustering, are especially challenging without holistic in-context processing. We see quantitative evidence of such decomposition resistance in our experimental results, presented in \Cref{sec:experiments-popmedqa}. 

PopMedQA presents realistic questions from population health. Whereas clinical medicine and biomedicine focus on interventions upon individual patients, population health concerns questions about groups of patients. PopMedQA's questions were reviewed by both clinicians and population health experts for pertinence and validity. 

\subsection{PopMedQA Pipeline}
\label{sec:popmedqa_pipeline}

Since the questions in PopMedQA are very long, they are generally not possible for humans (or computers) to solve or verify. Thus, a correct pipeline cannot generate questions and then (independently) generate labels. To ensure answers are correct, the questions and the data involved in the answers must be jointly generated.  The overall pipeline is as follows.

\paragraph{1. Task Definition} This involves specifying the schema of the answer (e.g. a list of lists of patient IDs) as well as the scoring function between the answer and the truth (see Appendix~\ref{appx:task_scoring} for details on each task's metric). It also involves specifying the cohorts that should be generated so that the answer can be synthesized from them. For example, planted clique is the task of finding the most cohesive or similar size-$k$ subset of patients. The task indicates two cohorts should be created: the size-$k$ clique, and the $N-k$ remaining patients. 

\paragraph{2. Question Ideation} Given a task, abstract question ideas (or template) are generated. The idea does not have a concrete question statement, nor does it have individual patients generated. Instead, the idea has a question template parameterized by $N$, e.g. ``among these {{N}} ED and urgent care records from the last 72 hours in our metro area, is there a subset of patients presenting with an unusual combination of symptoms...''. The idea also gives question-specific descriptions of the cohorts that need to be generated. For example: 
\begin{enumerate}
    \item Syndromic subset of size $\min(12, \mathrm{int}(N/4))$: a subset of patients from recent ED and urgent care records in a specific zip code over the last 72 hours. All have a documented chief complaint that includes a combination of 'severe gastrointestinal distress (vomiting/diarrhea)' AND 'unusual rash'.
    \item Other population of size $N - \mathrm{min}(12, \mathrm{int}(N/4))$: all other ED and urgent care records from the last 72 hours, showing typical presentations like chest pain, respiratory infections, and minor trauma, without the combination of severe GI distress and rash.
\end{enumerate}

\paragraph{3. Question Sampling} For each question idea, cohorts are generated at a large size $N_{\textrm{max}}$. To create concrete questions of size $N$, the idea's cohorts are subsampled at the specified value of $N$. 

\paragraph{4. Clinician Validation} To ensure the quality and practical relevance of the benchmark, questions undergo a clinician validation step. Clinician reviewers (M.D.s) score each question from 1--10 on three metrics: (1) \emph{Realism} (how likely or commonly the questions would arise in practice), (2) \emph{Difficulty} (of solving them by hand), and (3) \emph{Coherence} (whether the cohorts are well-defined and distinct). Questions are retained in PopMedQA only if they score at least 5 on all three metrics.

\section{\texttt{neopatient}: Language-Controlled Generation of Patient Records}
\label{sec:neopatient}

\begin{figure*}[t!]
  \centering
  \includegraphics[width=0.98\linewidth]{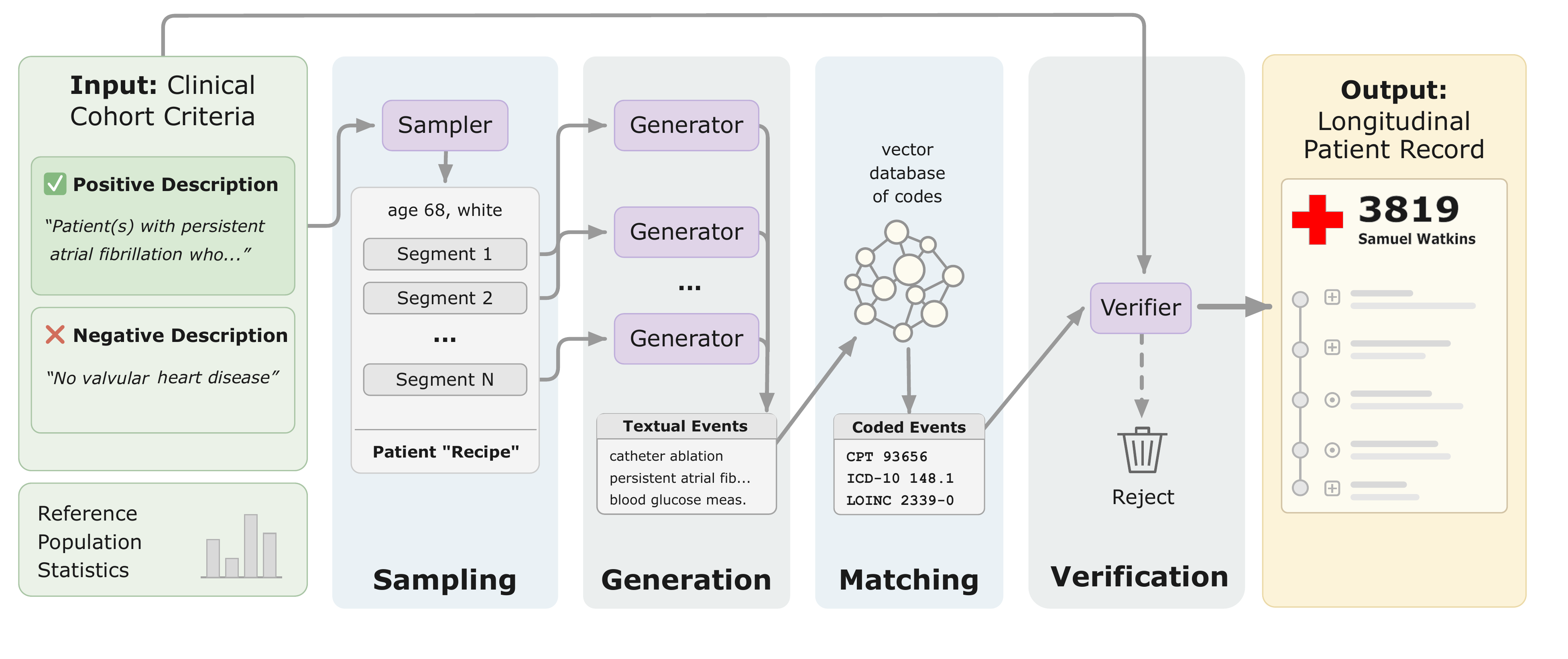}
  \caption{The neopatient architecture for language-controlled artificial patient generation. The pipeline transforms
  natural language criteria describing a cohort into a set of coded longitudinal records in the MEDS format. The architecture consists of four primary stages: (1) Sampling, where an LLM generates a ``patient recipe'' that defines
  demographics and divides the patient's life trajectory into discrete temporal segments to ensure coherence; (2) Generation, where
  longitudinal medical events are produced for each segment; (3) Matching, where a vector database maps natural language descriptions
  to standardized medical codes; and (4) Verification, a final correctness check where an LLM validates that the completed record
  strictly satisfies the original cohort specification.}
  \label{fig:neopatient_arch_appendix}
\end{figure*}

To generate the large volume of synthetic data required for PopMedQA, we implemented a new software library, neopatient, for language-controlled generation of artificial patient records. Unlike rule-based generators (like Synthea \citep{walonoski2018synthea}), patient trajectories in neopatient are controlled through natural language descriptions, allowing for the creation of complex clinical cohorts without the need for custom simulation code or state machines. 

Generating longitudinal patient records with LLMs presents several inherent challenges. First, \emph{output length} constraints are significant; LLM generations are typically limited to less than 64K tokens, and they often become unreliable when producing long, structured outputs at that scale. Second, \emph{coding knowledge} is limited, as LLMs do not precisely know the vast and frequently updated medical ontologies. Third, maintaining \emph{clinical plausibility} requires managing complex sequential dependencies, such as ensuring a prescription follows an appropriate diagnosis. Finally, the need for \emph{batching for efficiency} when generating large cohorts of hundreds of patients restricts the use of complex, multi-turn agentic loops. The neopatient pipeline consists of four primary stages designed to address these challenges:

\paragraph{1. Sampling} An LLM generates individualized ``patient recipes'' that define demographics and divide the patient's life trajectory into discrete temporal segments. This high-level blueprint ensures long-term clinical coherence—such as maintaining consistent medication dosages—while segmentation allows the system to bypass LLM context limits and avoid the ``drifting'' common in long-form generation.

\paragraph{2. Generation} For each recipe, an LLM produces longitudinal medical events across the temporal segments. For each event, the LLM generates a natural language description as well as a target coding system (e.g., ICD-10 or SNOMED). At this stage, the descriptions are medically plausible but may not yet match official ontology strings exactly.

\paragraph{3. Matching} Because LLMs are prone to hallucinating invalid codes or using imprecise language, the system uses a precomputed vector database (e.g., ChromaDB) to map free-text descriptions to standardized medical codes (SNOMED, ICD-10, LOINC, RxNorm, etc.).

\paragraph{4. Verification} A final correctness check is performed where an LLM validates each completed record against the original input specifications. This acts as an automated quality gate, filtering out records that failed to follow the recipe or accidentally triggered exclusion criteria.

The resulting records are produced in the Medical Event Data Standard (MEDS) format \citep{meds2024iclr}. neopatient is designed to be scalable, using LLM batch APIs and state-tracking for resumability, enabling the cost-effective generation of tens of thousands of records. 

\paragraph{Static Resources} A vector database of medical codes ensures that natural language descriptions are accurately mapped to standardized ontologies. This database was constructed by embedding code descriptions (using Qwen 3 8B to produce 4096-dimensional vectors). Second, the library maintains a set of reference statistics derived from real-world EHR data. these statistics define the typical length and density of both inpatient and outpatient longitudinal records, ensuring that the generated synthetic trajectories reflect realistic clinical patterns.

\paragraph{Note on Realism} The primary goal of neopatient is to generate records that strictly adhere to provided clinical specifications rather than to exhaustively replicate every facet of real-world patient records. This design priority serves the core objective of PopMedQA: to isolate and evaluate the specific computational challenges of the verbose context problem in a controlled setting.

Nonetheless, language-controlled generation potentially enables researchers to maximize realism while preserving privacy. The neopatient prompt can be optimized end-to-end (with TextGrad \citep{yuksekgonul2025textgrad}, GEPA \citep{agrawal2025gepa}, or similar) to generate records that maintain high fidelity with real, sensitive patient records. The resulting prompt can be more easily reviewed for privacy compliance than models trained with deep learning, which encode sensitive patient information in their weights. 

\vspace{-1mm}
\section{Experiments}
\label{sec:experiments-popmedqa}

We conduct a thorough evaluation of a range of language models on PopMedQA, testing models that range from 7B parameters to frontier scale, with context lengths from 128K to 2M tokens. On top of baseline models, we examine four families of interventions or ablations designed to improve long-context performance:

\vspace{-1mm}
\paragraph{Prompting strategies} We consider three ways of representing patient records as text: (1) a baseline method that replaces codes with truncated descriptions, (2) a naive method using only raw codes, and (3) a "codebook" method that, at the beginning of each prompt, maps unique codes to IDs, thereby reducing redundancy (see \Cref{appxfig:prompting,appxfig:prompting_codebook}).

\vspace{-0.5mm}
\paragraph{Prompt compression} We evaluate two methods: rendering text to images using the Glyph pipeline~\cite{cheng2025glyph} and LLMLingua-2 text chunk compression using the \nolinkurl{microsoft/llmlingua-2-xlm-roberta-large-meetingbank} model~\cite{jiang2023llmlingua}.

\vspace{-0.5mm}
\paragraph{Reasoning} We expand inference-time compute using chain-of-thought prompting~\cite{wei2022chain}. "Think" models were run with default temperatures and no upper bound on answer length, while non-reasoning models were run with temperature 0 and a 2048 token limit.

\vspace{-0.5mm}
\paragraph{Multi-turn interaction} We utilize agentic context engineering and decomposition via systems like MARS, LongCEPO, and Claude Code. These systems were allowed to perform multiple queries to solve a single question, with Claude Code having a \$5 USD limit per answer.

\subsection{Results and Discussion}
\label{sec:results}

\begin{figure*}[t!]
\centering
\includegraphics[width=\linewidth]{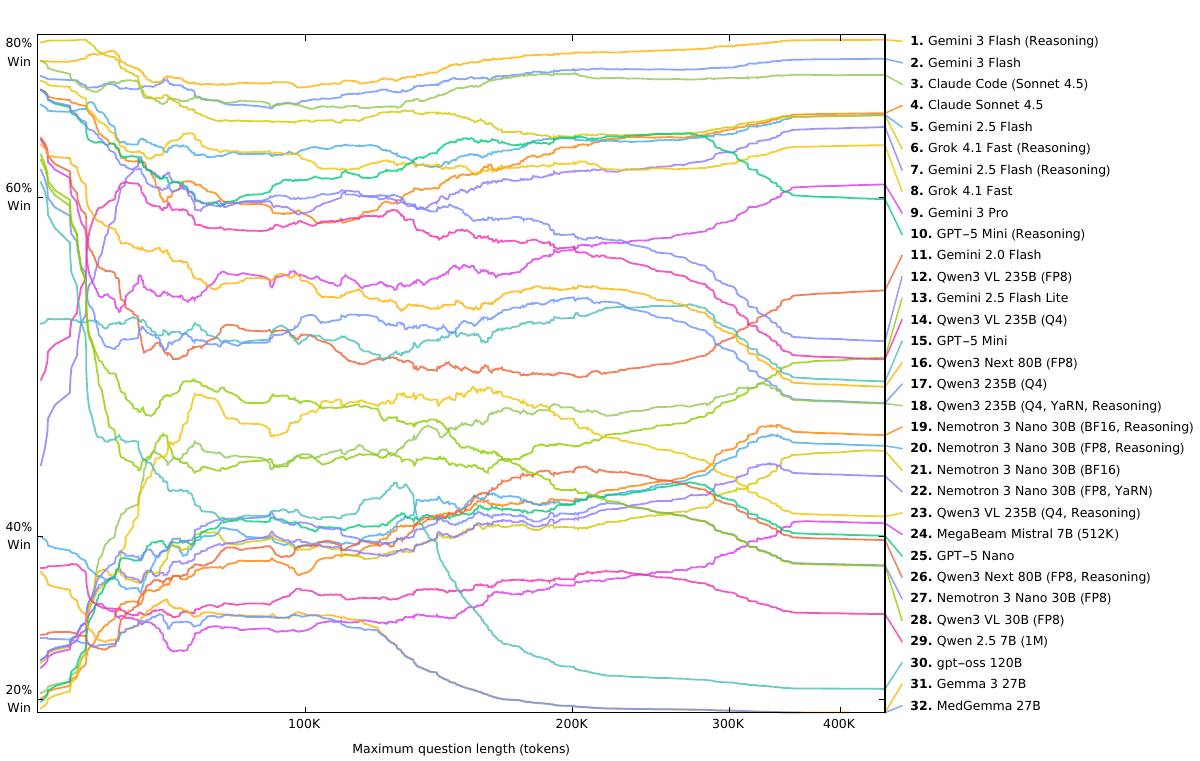}
\caption{Leaderboard performance of models across all tasks on PopMedQA. The y-axis is the percentage of comparisons the model won against all other models. The x-axis restricts these comparisons to questions up to a given token length. We observe that performance stresses frontier-level capabilities.}
\label{fig:leaderboard}
\end{figure*}

The primary experimental results on PopMedQA are presented in \Cref{fig:leaderboard,fig:meta_analysis_appendix}. Overall, the leaderboard results align with expectations, as frontier models demonstrate superior performance, followed by large open-source models with strong long-context capabilities. The tasks in PopMedQA effectively stress frontier-level capabilities, with performance declining as context length increases. Detailed task-specific scoreboards are provided in \Cref{fig:full_results_appendix}.

\paragraph{Clinical competence and long-context capability are required.} Maintaining a high rank at increased context lengths is not guaranteed. While models like Gemini 3 Flash are consistent, smaller models like Nemotron 3 Nano only rise in rank as the context expands, indicating that both medical domain knowledge and long-context processing are essential for PopMedQA.
\vspace{-1mm}
\paragraph{Generic prompt compression is fragile.} Compressed prompts, such as those generated by rendering text to images, significantly degrade model performance, particularly in instruction following. Domain-independent compression methods appear to strip away critical information needed for complex medical reasoning.
\vspace{-1mm}
\paragraph{Medical pretraining does not substantially improve performance.} General-purpose frontier models often outperformed specialized medical models on PopMedQA. This suggests that the primary challenge is not a lack of clinical knowledge but rather the ability to perform robust, multi-hop reasoning over the verbose contexts characteristic of longitudinal EHR data. Other works have cast doubt on the utility of medically-specialized language models~\citep{jeong2024medical}.

\paragraph{Agentic (multi-turn) decomposition is ineffective and/or cost-prohibitive.} While agentic systems showed relative strength, they failed to achieve absolute performance gains that justify their high computational and financial costs. Systems like MARS and LongCEPO frequently harmed instruction-following and absolute performance compared to monolithic baselines. Furthermore, Claude Code did not surpass the efficiency of frontier monolithic models. This confirms that PopMedQA's tasks are resistant to simple decomposition and require holistic in-context reasoning.

\section{Conclusion}

The verbose context problem inhibits population-level EHR reasoning. Our results demonstrate that domain-independent methods—including prompt compression and agentic decomposition—fail to alleviate performance degradation. These findings reveal a significant unrealized opportunity to exploit domain-specific structure in language model inputs to enable robust population-scale reasoning. Beyond the verbose context problem, language-controlled generation of artificial patient records, as in neopatient, can accelerate AI research in population health.

\bibliography{workshop_paper}
\bibliographystyle{icml2026}

\newpage
\appendix
\onecolumn

\section{PopMedQA Details}
\label{appx:popmed_details}

\begin{figure}[H]
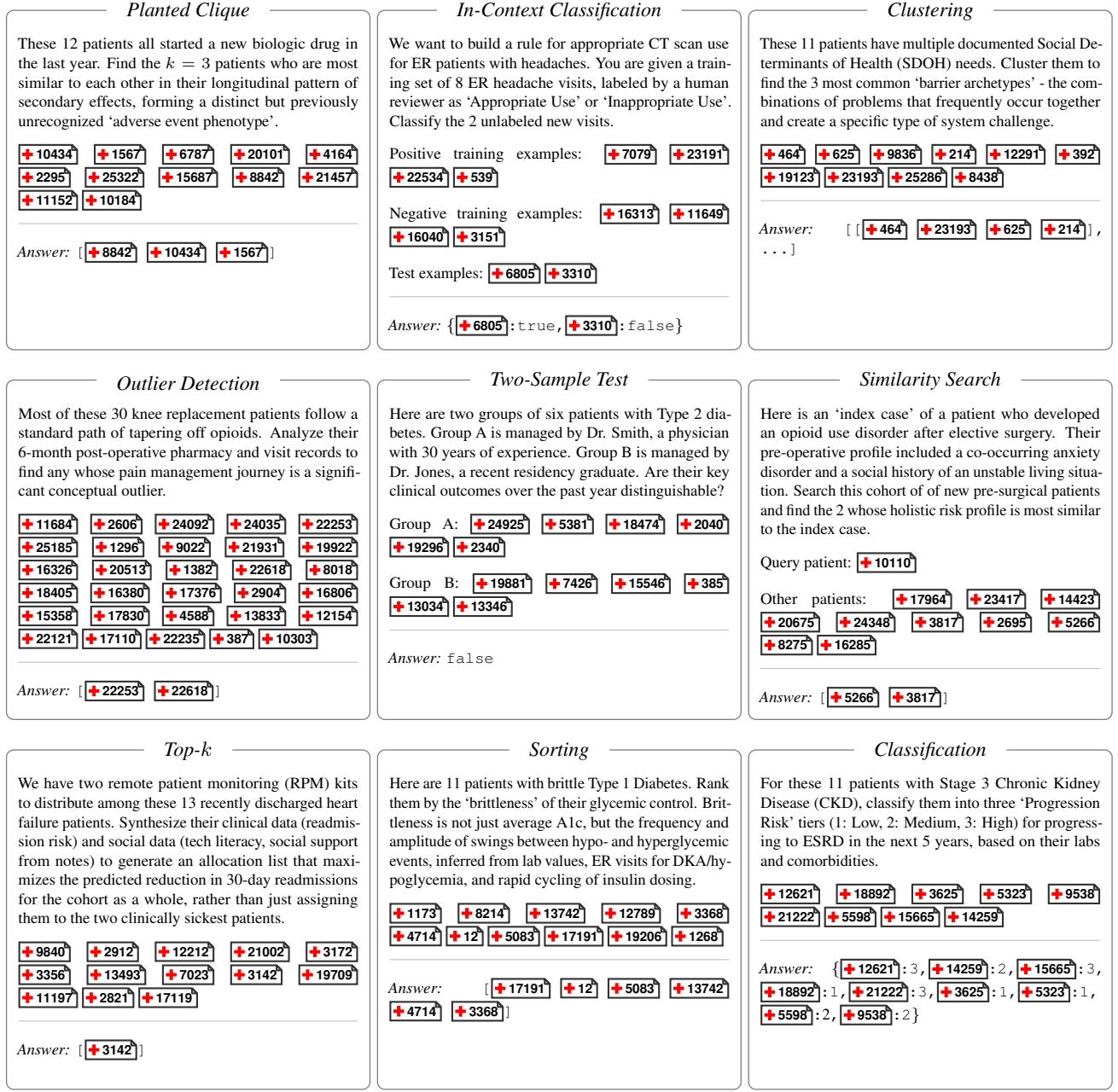

\noindent
\begin{tcbraster}[raster columns=3, raster width=\linewidth, raster equal height, raster column skip=1mm, raster force size]
\begin{questioncard}[Planted Clique]
\input{questions/ace402bc-892d-4dba-8115-21d0fe4c3667.tex}
\end{questioncard} 
\begin{questioncard}[In-Context Classification]
\input{questions/fe2739bf-5aa0-41cc-a13f-1c93a7b779f1.tex}
\end{questioncard}
\begin{questioncard}[Clustering]
\input{questions/d39e9d19-3661-48d3-97fe-3fb1c4f57ec4.tex}
\end{questioncard}
\begin{questioncard}[Outlier Detection]
\input{questions/c5bf34f8-432a-414e-a1df-c8a93b95fb6e.tex}
\end{questioncard}
\begin{questioncard}[Two-Sample Test]
\input{questions/3443cae0-391c-4e52-b10a-29f58a746456.tex}
\end{questioncard}
\begin{questioncard}[Similarity Search]
\input{questions/d812316a-64ab-4e47-9230-e91170c78962.tex}
\end{questioncard}
\begin{questioncard}[Top-$k$]
\input{questions/d3b1ae9a-0bbe-4a6a-930c-38b6ea9d5d31.tex}
\end{questioncard}
\begin{questioncard}[Sorting]
\input{questions/b348896d-55dc-49c1-95f4-4a4bb05b5817.tex}
\end{questioncard}
\begin{questioncard}[Classification]
\input{questions/e196b380-08ee-4f1a-b8c9-2b3dc473ab17.tex}
\end{questioncard}
\end{tcbraster}
\caption{Example questions from PopMedQA. Each question poses one of nine computational tasks. Each patient record is visually depicted as a boxed patient ID number.}
\label{fig:popmed_all_tasks}
\end{figure}

\newpage

\begin{figure*}[ht]
\centering
\includegraphics[width=\linewidth]{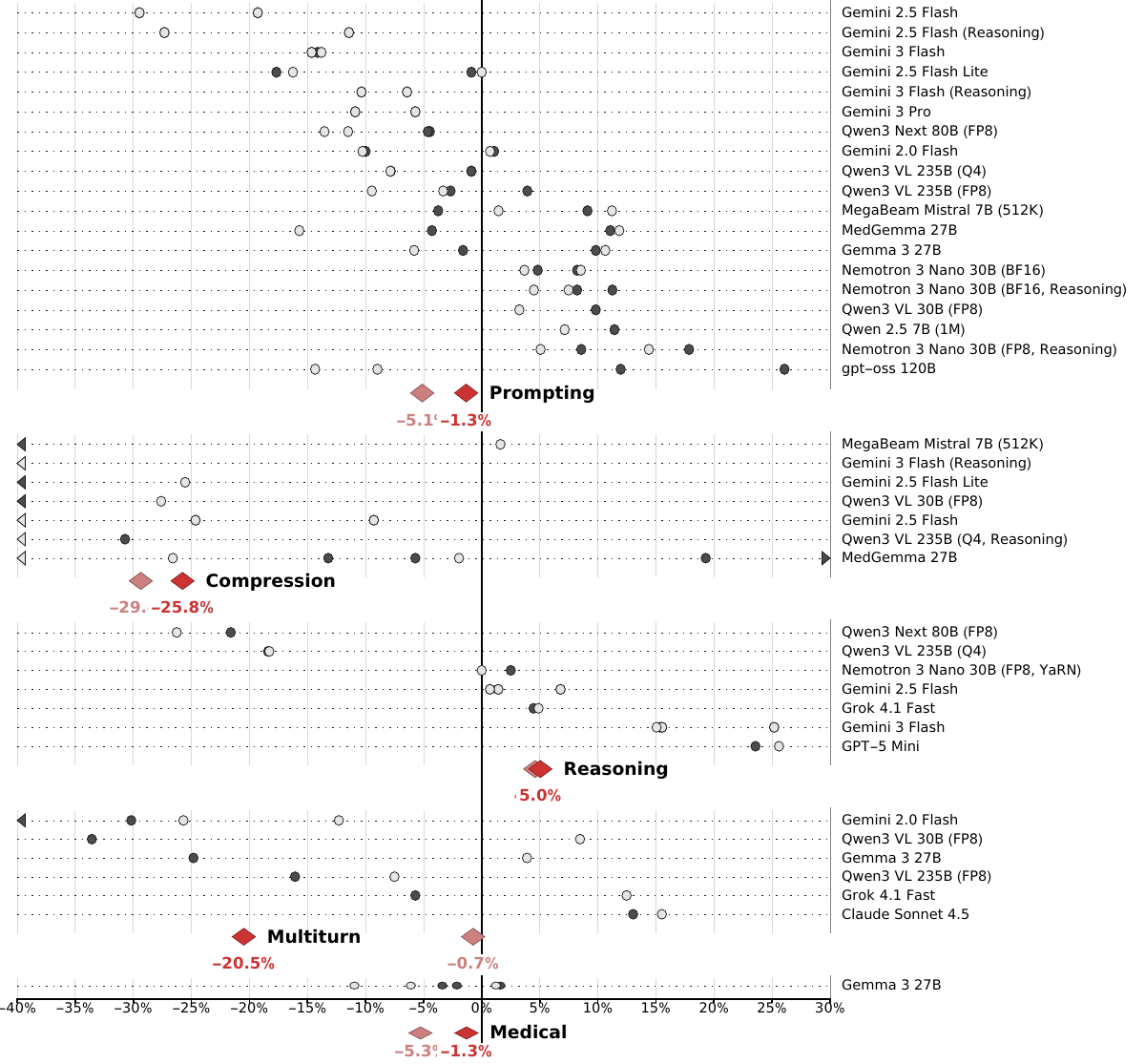}
\caption{Meta-analysis of ablations on PopMedQA. Each dot compares two runs on PopMedQA: a baseline and an ablation. The model's name is on the right; different families of ablations are presented. A dot's position quantifies the effect of the ablation. A dark dot at -5\% indicates that the baseline model won 55\% of head-to-head comparisons, and therefore the ablation had a negative effect. A light dot restricts the scoring to examples where both models gave an answer; this distinction is important when the ablation affects the model's capability to return correctly-formatted answers. For each family, the mean ablation effects are shown as diamonds. Overall, we find that most families of ablations, besides reasoning, are ineffective on PopMedQA.}
\label{fig:meta_analysis_appendix}
\end{figure*}

\newpage

\begin{figure}[htbp]
\centering
\begin{minipage}{0.30\textwidth}
    \centering
    \includegraphics[width=\textwidth]{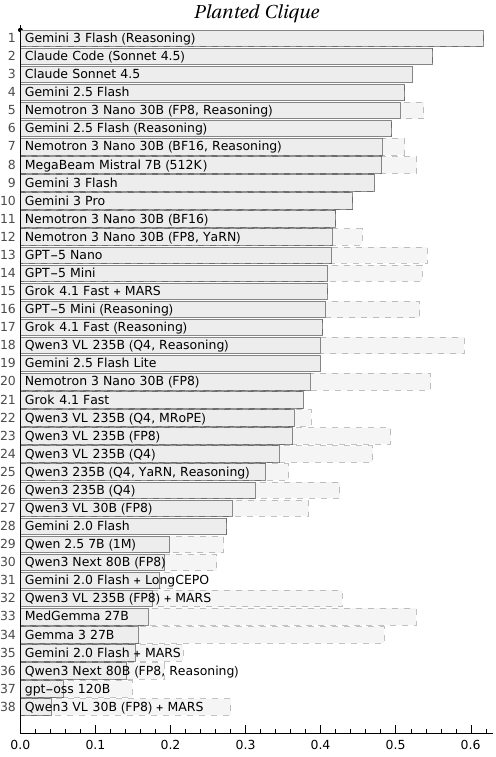}
\end{minipage}\hfill
\begin{minipage}{0.30\textwidth}
    \centering
    \includegraphics[width=\textwidth]{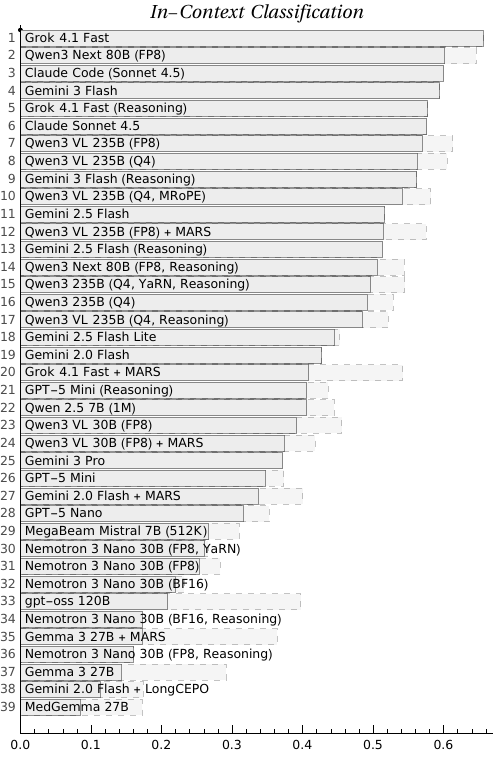}
\end{minipage}\hfill
\begin{minipage}{0.30\textwidth}
    \centering
    \includegraphics[width=\textwidth]{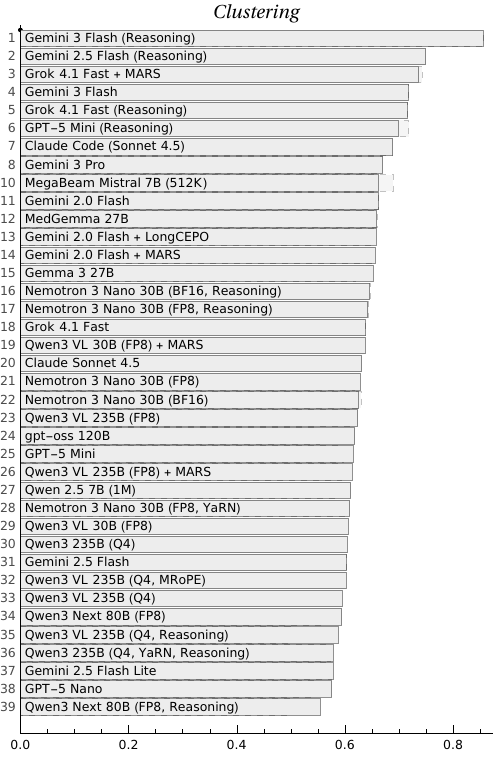}
\end{minipage}
\vspace{0.1cm}
\begin{minipage}{0.30\textwidth}
    \centering
    \includegraphics[width=\textwidth]{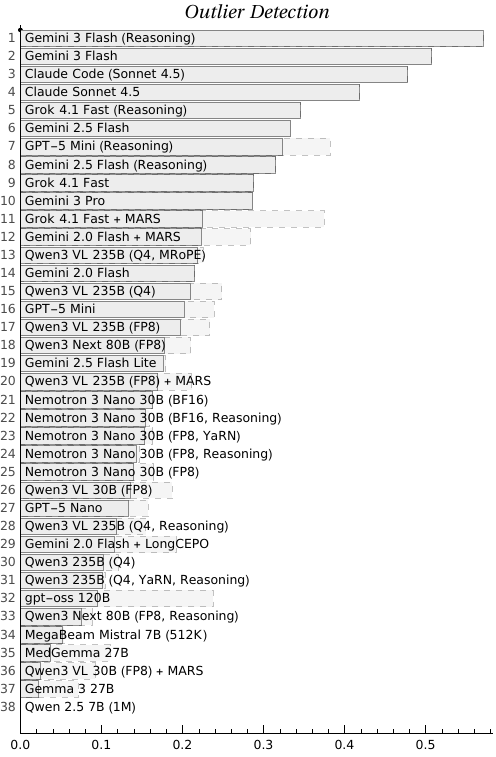}
\end{minipage}\hfill
\begin{minipage}{0.30\textwidth}
    \centering
    \includegraphics[width=\textwidth]{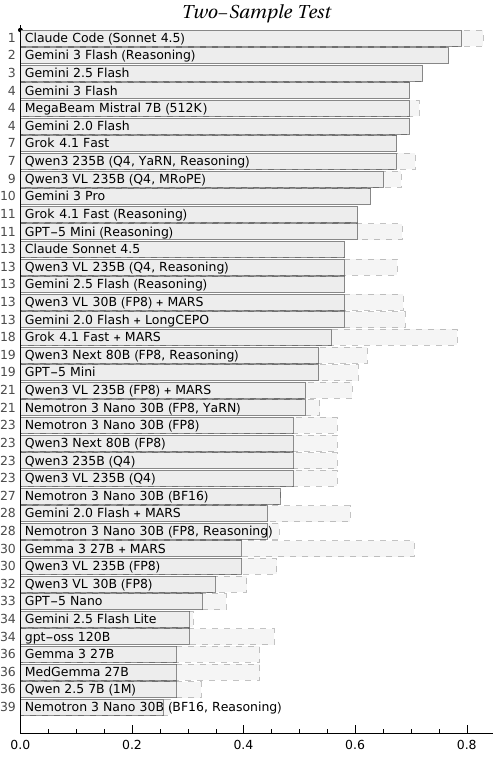}
\end{minipage}\hfill
\begin{minipage}{0.30\textwidth}
    \centering
    \includegraphics[width=\textwidth]{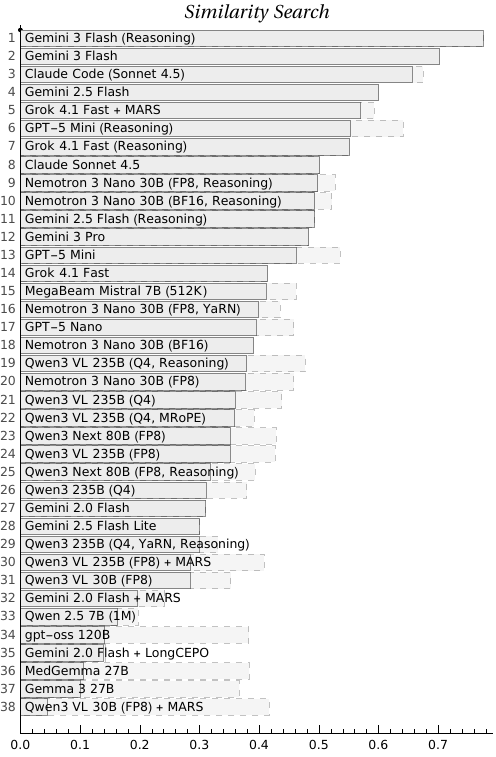}
\end{minipage}
\vspace{0.1cm}
\begin{minipage}{0.30\textwidth}
    \centering
    \includegraphics[width=\textwidth]{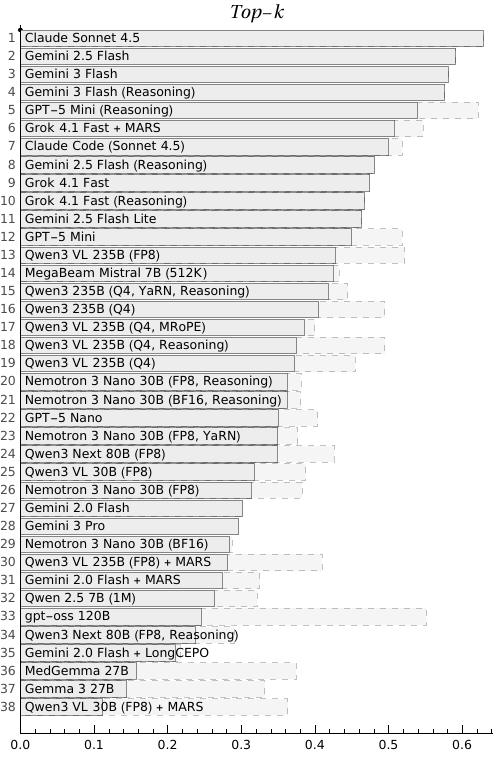}
\end{minipage}\hfill
\begin{minipage}{0.30\textwidth}
    \centering
    \includegraphics[width=\textwidth]{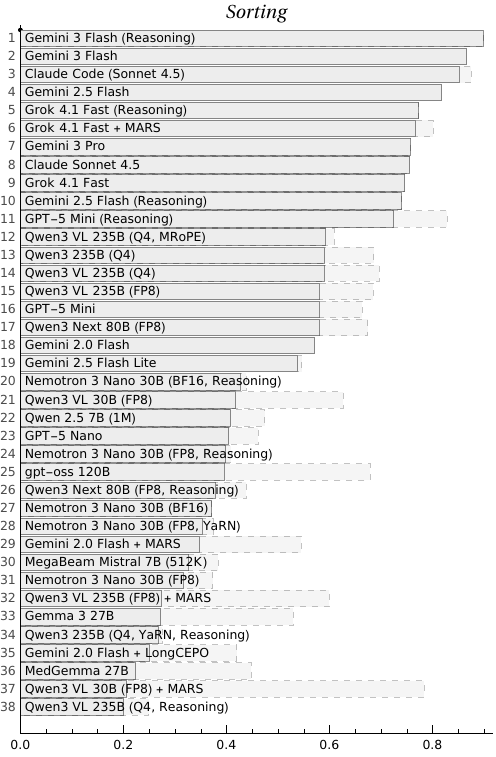}
\end{minipage}\hfill
\begin{minipage}{0.30\textwidth}
    \centering
    \includegraphics[width=\textwidth]{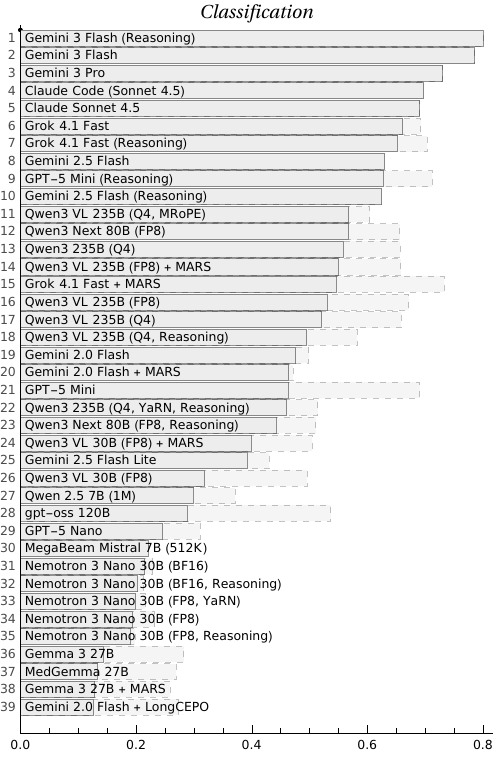}
\end{minipage}
\caption{Task-specific scoreboards. The y-axis indicates the model's rank, and the x-axis denotes its mean score on the task. The filled-in bars with solid borders indicate the mean over all the questions in the task. The faint bars with dashed borders indicate the mean over just the questions the model answered properly (i.e. in the correct format).}
\label{fig:full_results_appendix}
\end{figure}

\newpage

\begin{figure}[h!]
\centering
\begin{minipage}[t]{0.62\linewidth}
\promptexample{basic}{1.0}
\end{minipage}
\hfill
\begin{minipage}[t]{0.34\linewidth}
\promptexample{naive}{1.0}
\end{minipage}
\caption{Different prompting methods on the same patient record. (Left): the standard prompting method used as a baseline in this paper. It replaces the code altogether (since those are often not recognized by language models) by a truncated description of the code. (Right): a less verbose (and less informative) prompting method which includes the code but not the description.}
\label{appxfig:prompting}
\end{figure}

\begin{figure}[h!]
\centering
\promptexample{codebook}{0.93}
\caption{An alternative prompting method which attempts to eliminate redundancy across groups of patient records. It defines a succinct ID numbers for all unique codes (across all patients), and then references those IDs within the subsequent patient records.}
\label{appxfig:prompting_codebook}
\end{figure}

\clearpage

\section{Formalization of the Verbose Context Problem}
\label{sec:verbosecontext}

We consider heterogeneous prompts $x : \mathrm{list}(\mathrm{string} + \mathcal{C})$ for LLMs. $\mathcal{C}$ is a set of concepts distinguished from the other unstructured parts of the prompt. As the canonical example in this paper, we take $\mathcal{C}$ to be the set of all medical codes. The verbose context problem arises when converting $c \in \mathcal{C}$ to strings. Let $\mathrm{str}(c)$ be a string which conveys $c$ to a target language model with the desired level of precision. For a medical code, it may not be sufficient to pass just the abbreviated code; instead, much of the extended description may be needed as well. To be processed by the language model, the string must be tokenized, and then each token's embedding is looked up in the embedding matrix. This produces a variable-length sequence of vectors $\mathrm{emb}\circ \mathrm{tok}\circ\mathrm{str}(c) \in \mathrm{list}\ \mathcal{E}$, where $\mathcal{E}$ is the LLM embedding space, which is typically a few thousand dimensions. Given a probability distribution over prompts $x$, the baseline (overall) expected context length is $N$. $M$ is the context length that originates from $\mathcal{C}$.
\begin{align*}
M &= \mathbb{E}_x\ \sum_{c \in x \cap \mathcal{C}} \textrm{Length}(\mathrm{emb} \circ \mathrm{tok} \circ \mathrm{str}(c)) \\
N &= \mathbb{E}_x\ \textrm{Length}(\mathrm{emb} \circ \mathrm{tok} \circ \mathrm{str}(x))
\end{align*}
(In the second line, we slightly abuse notation to have $\text{str}(x)$ operate on all parts of the prompt). Informally, the verbose context problem is that $N$ is too long to be practical.
\begin{definition}[Verbose Context Problem]
This occurs when $M / N$ is large, i.e. $\Omega(1)$ as $N \to \infty$.
\end{definition}

\vspace{0.4cm}

\section{Task Scoring}
\label{appx:task_scoring}

\begingroup
\footnotesize
\begin{longtable}{p{3.6cm}p{2.0cm}p{8.5cm}}
\toprule
\textbf{Task} & \textbf{Metric} & \textbf{Details} \\ \midrule
\endfirsthead
\toprule
\textbf{Task} & \textbf{Metric} & \textbf{Details} \\ \midrule
\endhead
\bottomrule
\endfoot
Planted Clique & Precision & $|\text{Predicted} \cap \text{True}| / k$ \\ \midrule
In-Context Classification & Accuracy & Binary classification accuracy on the test set \\ \midrule
Clustering & Rand Index & Proportion of correctly identified pairs (same vs.\ different) \\ \midrule
Inverse-Propensity Weighting & Absolute Error & Absolute difference between predicted and true ATE (Hajek) \\ \midrule
Outlier Detection & $F_1$ Score & Harmonic mean of precision and recall \\ \midrule
Two-Sample Test & Accuracy & 0--1 accuracy in identifying if distributions differ \\ \midrule
Similarity Search & Precision@$k$ & Proportion of model's top-$k$ that are in the true top-$k$ \\ \midrule
Top-$k$ & Precision@$k$ & Proportion of model's top-$k$ that are in the true top-$k$ \\ \midrule
Sorting & Kendall's $\tau$ & Restricted to pairs from different ground-truth cohorts \\ \midrule
Classification & Accuracy & Multi-class accuracy across all categories \\
\end{longtable}
\endgroup

\newpage

\section{Related Work}
\label{appx:related_work}

\paragraph{Long-Context Benchmarks.} Since the advent of modern language models, dozens of long-context benchmarks have been developed \citep{tay2021long, bai2023longbench, zhang2024infbench, yen2025helmet, bai2025longbench}. Context lengths under evaluation have increased from 4K to above 1M. Different evaluations attempt to isolate different failure modes. Early diagnostic tests focus on retrieval failures and positional bias (e.g. ``lost in the middle''). Newer evaluations target reasoning degradation in multi-step tasks \citep{openai2025graphwalks}. In these benchmarks, difficulty and context length are driven primarily by (1) the amount and density of irrelevant distractors, and (2) the number of reasoning hops required to bridge dispersed information \citep{vodrahalli2024michelangelo, openai2025mrcr}. PopMedQA emphasizes a different cause of context bloat (verbosity) in order to expose different failure modes.  

While some benchmarks isolate long-context reasoning through abstract tasks~\citep{openai2025graphwalks}, others prioritize naturalistic, real-world inquiries~\citep{bai2023longbench,artificialanalysis2025lcr}. Our work belongs to the latter category. We encourage more situated, concrete study of long-context problems by recognizing and addressing their domain-specific aspects. We aim to further close the gap between benchmark performance and real-world utility. 

\paragraph{AI on EHRs.} Existing benchmarks for AI in Electronic Health Records (EHRs) evaluate a wide range of clinical and administrative capabilities.
  For structured data, EHRSHOT \citep{wornow2023ehrshot} and INSPECT \citep{huang2023inspect} assess few-shot clinical prediction and
  algorithmic fairness within individual patient timelines. EHRSQL \citep{lee2022ehrsql} and MedAgentBench
  \citep{jiang2025medagentbench} extend these evaluations to cohort-level queries; however, these frameworks primarily test the
  model’s ability to translate natural language into structured queries (SQL or FHIR) that delegate computational aggregation to an
  external database engine. For unstructured text, MedAlign \citep{fleming2024medalign} and MedFactEval \citep{grolleau2026medfacteval}
  focus on instruction-following and factuality within clinical notes. PopMedQA diverges from these approaches by shifting the
  analytical focus to population health, requiring models to perform holistic, in-context reasoning across the raw longitudinal
  records of cohorts of 10 to 50 patients simultaneously. This framework unlocks complex use cases in population-level pattern
  discovery, such as identifying latent clinical clusters or detecting sparse anomalies across disparate patient trajectories, that
  cannot be readily addressed by standard query-based aggregation or individual-level processing.

\paragraph{Population Health Analytics.} Population health analytics focuses on the health outcomes of groups of individuals and the distribution of these outcomes within the group \citep{kindig2003population}. Its primary objectives are to quantify disease burden and guide resource allocation to ensure equitable and efficient healthcare delivery. To achieve this, established risk adjustment systems like the Johns Hopkins Adjusted Clinical Group (ACG) System \citep{weiner2012acg}, the CMS Hierarchical Condition Category (CMS-HCC) model \citep{pope2004risk}, and comorbidity indices such as Charlson \citep{charlson1987new} and Elixhauser \citep{elixhauser1998comorbidity} are widely employed. These tools primarily rely on rule-based aggregation of structured diagnosis codes and pharmacy data to perform retrospective financial risk stratification and predict healthcare utilization. However, these statistical frameworks are often limited by fragile or manual feature engineering that cannot capture the complex dependencies within a patient's history, leading to low individual-level predictive accuracy (e.g., $R^2$ values frequently below 15\% for prospective cost prediction) \citep{pope2004risk}. PopMedQA evaluates a more flexible, yet still code-centric, alternative where modern language models perform comprehensive longitudinal reasoning over groups of patient records across entire cohorts.

\paragraph{Alternative Concept Representations.} Are there more succinct ways to convey concepts to language models than language itself? Multiple lines of work support this general idea. The common strategy is to inject concepts as vectors at different model layers, in lieu of providing more input text. In prefix tuning \citep{li2021prefix}, the output embeddings of the earlier part of a prompt are truncated and directly optimized to improve the accuracy of subsequent generation. Cartridges \citep{eyuboglu2025cartridges} also condense the prefix by optimizing the contents of key-value caches in attention layers. Steering vectors \citep{jahanian2020steerability, subramani2022extracting} are added to activations to control generation in an input-agnostic manner. Rendering text to images and using vision language models can be effective for long-context inference \citep{zheng2024multimodal, cheng2025glyph, wei2025deepseek}.

\end{document}